\documentclass[11pt,a4paper]{article}
\usepackage[hyperref]{acl2020}
\usepackage{times}
\usepackage{latexsym}
\usepackage{amssymb}
\usepackage[export]{adjustbox}
\usepackage{tabularx}
    \newcolumntype{C}{>{\centering\arraybackslash}X}

\usepackage{microtype}
\usepackage[bottom]{footmisc}

\aclfinalcopy % Uncomment this line for the final submission

\definecolor{britishracinggreen}{rgb}{0.0, 0.5, 0.0}

\title{A Study on Efficiency, Accuracy and Document Structure \\ for Answer Sentence Selection}

\author{Daniele Bonadiman \\
  University of Trento \\
  Trento, Italy \\
  \texttt{d.bonadiman@unitn.it} \\\And
  Alessandro Moschitti \\
  Amazon Alexa \\
  Manhattan Beach, CA, USA \\
  \texttt{amosch@amazon.com} \\}

\date{}

\begin{document}
\maketitle
\begin{abstract}
An essential task of most Question Answering (QA) systems is to re-rank the set of answer candidates, i.e., Answer Sentence Selection (A2S). These candidates are typically sentences either extracted from one or more documents preserving their natural order or retrieved by a search engine. Most state-of-the-art approaches to the task use huge neural models, such as BERT, or complex attentive architectures.
In this paper, we argue that by exploiting the intrinsic structure of the original rank together with an effective word-relatedness encoder, we can achieve competitive results with respect to the state of the art while retaining high efficiency. Our model takes 9.5 seconds to train on the WikiQA dataset, i.e., very fast in comparison with the $\sim 18$ minutes required by a standard BERT-base fine-tuning. 

\end{abstract}

\section{Introduction}

%Motivation for QA
In recent years, there has been a renewed interest in Question Answering (QA) led by both industrial needs, e.g., the development of personal assistants such as Google Home, Alexa, Siri, as well as academic research on neural networks. Regarding the latter, answer sentence selection/re-ranking (A2S) \cite{wang2007jeopardy,yang2015wikiqa} and machine reading comprehension (MRC) \cite{richardson2013mctest,rajpurkar2016SQuAD} have been largely explored.

A2S consists of selecting sentences that can answer a target question from documents or paragraphs retrieved from the web by a search engine.

MRC regards the extraction of an exact text span from a document answering the question, where the document is usually provided with the target question. 

% Excessive claim
%Even though MRC is gaining more and more popularity, A2S is more suitable for a production scenario since a combination of a retrieval engine together with a sentence selector model constitutes a QA system. In contrast, MR has been mainly developed to find answers in a paragraph or a text of limited size. Several models have been proposed to adapt MR models to a retrieval scenario (i.e., applicable to retrieval results), e.g., \cite{DBLP:journals/corr/ChenFWB17,DBLP:journals/corr/abs-1906-04618,kratzwald-feuerriegel-2018-adaptive}. 
Even though MRC is gaining more and more popularity, A2S is more suitable for a production scenario since a combination of a retrieval engine together with an automatic sentence selector can already constitute a QA system.\footnote{For the aim of this paper we analyze QA models that operates on unstructured text only.} In contrast, MRC has been mainly developed to find answers in a paragraph or a text of limited size. 
Even though, several models have been proposed to adapt MRC to an end-to-end retrieval setting, e.g., \cite{chen2017reading,kratzwald2018adaptive},  the deployment of MRC system in production is challenged by two key factors: the lack of datasets for training MRC with realistic retrieval data, and the large volume of relevant content needed to be processed, i.e., MRC cannot efficiently process a large amount of retrieved data. 

In contrast, A2S research originated from the TREC competitions \cite{wang2007jeopardy}; thus, it has targeted large databases of unstructured text from the beginning of its development. Neural models have significantly contributed to A2S with new techniques, e.g.,  \cite{wang2016compare, qiao2019understanding, nogueira2019passage}. 
Recently, new approaches for pre-training neural language models on large amount of data, e.g., ELMO \cite{Peters:2018}, GPT \cite{radford2019language}, BERT \cite{devlin2019bert}, RoBERTa  \cite{liu2019roberta},  XLNet \cite{yang2019xlnet}, have led to major advancements in several NLP subfields. These pre-training techniques allow for creating models that automatically capture dependencies between the sentence compounds. Interestingly, the resulting models can be easily adapted to different tasks by simply fine-tuning them on the target training data.

Unfortunately, all the models above, (especially the Transformer-based architectures) require many layers and a considerable number of parameters (up to 340 million for BERT Large). This poses three critical challenges for having such models in production: firstly, it requires powerful GPUs to achieve an acceptable service latency. Secondly, although the classification of candidates can be parallelized, the required number of GPUs will prohibitively increase the operational cost. This also has a huge environmental issue, as pointed out by \citet{strubell2019energy}.
Last, transformer-based architectures require many resources for pre-training, e.g., both data and compute power (TPUs). These resources may not be available for low resource languages or domain-specific applications.

%first, the latency of the approach highly increases with the number of candidates. 
%For example, to obtain a target Recall, the model has to process hundreds of candidates, which takes several seconds, also using powerful GPUs.
In this paper, we study and propose solutions to design accurate A2S models, still preserving high efficiency. We first note that (i) the primary source of inefficiency is, unfortunately, the contextual embedding, e.g.,  language models produced by Transformer or also previous methods such as ELMo. These introduce at least one order of magnitudes more of parameters in the A2S models. (ii) The other significant source of inefficiency is the attention mechanism.

As both of the above features critically impact accuracy, we provide an alternative to preserve it as much as possible. In particular, we model all candidates for a given question to capture the global structure of the document or the rank as provided in input to the model.
Our experiments verify the hypothesis that in several A2S datasets, the data often presents an underlying ranking structure. This refers not only to the relations between a question and all its candidates but also to the inter-dependencies among the candidates themselves. 

Our approach captures the above structure in the original rank.  For this purpose, we show that it is essential to implement two main logic blocks: (i) an encoder able to capture the relation between the question and each of its candidates, e.g., using an attention mechanism; and (ii) the structure of the sentences in the original rank, e.g., the similarities and dissimilarities between candidates. 
Regarding the second block, we use an additional layer constituted by a bidirectional recurrent neural network (BiRNN), which is fed with the representation of question/answer candidate pairs, where the latter are joint representations of the question and the answer, obtained by the question-answer encoder.

%For the first point, we capitalize on previous work on modeling attention, i.e., Compare-Aggregate (CA) framework \cite{wang2016compare}.  However, in contrast with the original model, our architecture performs the comparison and aggregation steps for both question and passage. Additionally, the proposed attention employs a gating mechanism to activate or not the alignment between the question and the passage. This is very useful to filter out noise due to the many possible alignments between a question with negative candidates.

Regarding the attention mechanism, we substitute it with a sort of \emph{static attention}, given by a cosine similarity between the embedding representation of the question and answer words. We show that this solution is very efficient and does not cause almost any drop with respect to the use of standard attention.

%The results derived on several datasets show that (i) we achieve state of the art among the efficient approaches; (ii) the lack of contextual embedding prevents the model from learning more complex function; thus, our joint model significantly improves the accuracy of efficient methods. (iii) our hard attention can replace the standard attention inefficient methods highly improve the speed of the approach.

The results derived on several datasets show that (i) we obtain better results than other efficient approaches; (ii) the lack of contextual embeddings prevents the model from learning more complex functions. We partially solve this problem by using our joint model, which significantly improves the accuracy of efficient methods. (iii) Our word-relatedness encoder can replace the standard attention to improve the speed of the approach.
%
%
%This allows us to use the question information both at a word-level and at passage-level. 
%
%Using a parallel model for both question and answer is common in state of the art models. 
%
%
%&For the second aspect, we 
%The idea is by performing listwise training of all the answer passages, using models that are sensitive to global ordering, e.g., Bidirectional LSTM (BiLSTM). 
%

We tested our models on four different datasets, the well-known WikiQA dataset, the adaptation of SQuAD \cite{rajpurkar2016SQuAD}, and Natural Questions \cite{47761} datasets to the A2S task. %, and  MSMarco \cite{nguyen2016ms}, a pure retrieval based dataset.
Additionally,  the results show that the global component outperforms the same models, not exploiting it.  Despite this adds a small overhead during training and testing, it is crucial to capture the structure of data. For example, the results on WikiQA show that BiRNN added to the Cosinet improves of $\sim$ 4 points previous baselines.
%Finally, we obtain improvement in all of the datasets using the structural component, suggesting that our model is effective in exploiting the structure of the sentences in the article or the rank. 

\section{A2S: Answer Sentence Selection}
\begin{table}[t]
\small
\renewcommand{\arraystretch}{1.1} 
\centering

\begin{tabularx}{\linewidth}{C} \hline {\textbf{How long was I Love Lucy on the air ?}}\\ \hline \hline
I Love Lucy is an American television sitcom starring Lucille Ball , Desi Arnaz , Vivian Vance , and William Frawley .\\  \hline

\leavevmode\color{britishracinggreen} The black-and-white series originally ran from October 15, 1951, to May 6, 1957, on the Columbia Broadcasting System (CBS).\\ \hline
After the series ended in 1957, however, a modified version continued for three more seasons with 13 one-hour specials, running from 1957 to 1960, known first as The Lucille Ball-Desi Arnaz Show and later in reruns as The Lucy–Desi Comedy Hour .   \\  \hline
I Love Lucy was the most watched show in the United States in four of its six seasons, and was the first to end its run at the top of the Nielsen ratings (an accomplishment later matched by The Andy Griffith Show and Seinfeld ).\\ \hline
I Love Lucy is still syndicated in dozens of languages across the world\\ \hline
\end{tabularx}
  \caption{\label{wikiqa:ex} An example of question/answer-candidate from WikiQA. In green the answer to the question.}
  \vspace{-1.em}
\end{table}

The task of Answer Sentence Selection (A2S)  can be formalized as follows: given a question q and a set of answer sentence candidates $C = \{c_1, c_2, ..., c_n\}$ the task is to assign a score $s_i$ for each candidate $c_i$ such that the sentence receiving the highest score is the one that most likely contains the answer. An excerpt of an A2S dataset is presented in Table \ref{wikiqa:ex}, in this example, the sentence that more likely answers the question ``\textit{How long was I Love Lucy on the air ?}'' is the second, i.e., ``\textit{The black-and-white series originally ran from October 15, 1951, to May 6, 1957, on the Columbia Broadcasting System (CBS).}''. In this scenario, the models need to assign a score to each sentence in such a way that the second sentence is ranked above the others. The task of A2S can be, in fact, effectively modeled as a re-ranking task.

Although re-ranking is a structured output problem, most state-of-the-art approaches treat the task of A2S as pointwise classification, i.e., classifying as positive sentences that contain the answer and as negative all the others. This design bias prevents A2S models from capturing the underlying structure of the original rank. However, in this paper, we argue that building systems capable of capturing such information is crucial for improving the performance of efficient A2S models.

\begin{table}[!t]
\centering
\resizebox{0.97\columnwidth}{!}{
\begin{tabular}{l|rrr}
\hline
                           & WikiQA          & SQuAD               & NQ-LA   \\ \hline\hline
\# questions (Q)              & 633             & 11873               & 6230            \\
\# sentences (C)                & 6165            & 63959                 & 193k          \\
\% Q answered   & 38.39 & 49.92 & 55.47 \\
avg. \# passages & 9.74            & 5.38              & 30.95           \\
avg. Q lenght       & 7.28            & 10.02    & 9.38            \\
avg. C lenght        & 25.36           & 23.75                & 98.76           \\
\hline
P@1 (random)              & 14.43         & 18.34           & 3.24       \\
MAP  (random)              & 25.15          & 43.81            & 12.33         \\
\hline
P@1 (RR)              & 46.09        & 30.54          & 46.06        \\
MAP (RR)              & 64.21         & 53.53              & 57.30          \\
\hline
P@1 (WO)              & 32.51         & 65.48       & 23.06        \\
MAP  (WO)              & 51.02          & 77.90             & 38.08         \\
\hline
P@1 (WO+RR)              &    56.38     & 73.12      &    41.01     \\
MAP  (WO+RR)              & 68.25          & 83.60           &53.98         \\
\hline

\end{tabular}
}
\caption{\label{tab:datasets} Statistics of the different datasets (the test-set are taken into account).}
\end{table}

\subsection{A2S Datasets}
\label{sec:datasets}
A2S datasets can be divided into two categories: retrieval based and document-based. The difference between the two categories resides in the source of the answer candidates. In the former, answer candidates are retrieved from a search engine, i.e., TrecQA \cite{wang2007jeopardy} and, more recently, MSMarco \cite{bajaj2016ms}. For the latter, a search engine is often used to retrieve the relevant document, but the task is to select the relevant answer candidate from the document itself. Notable examples of document-based A2S are the WikiQA dataset \cite{yang2015wikiqa} and the "long-answer" version of Natural Question \cite{kwiatkowski2019natural}. 

Despite the heterogeneous nature of the datasets, both types present strong features for detecting relevant answers in the candidate set: substantial lexical overlap between the question and the answer candidate and a global structure, i.e., the original order of the sentences in the rank.

\subsubsection{Lexical Overlap}
One of the strongest features in A2S datasets is the lexical overlap, i.e., whether words appear in both questions and answer candidates. The importance of this feature is highlighted in Table \ref{tab:datasets}. We used the number of unique words that appear in both the question and the candidates as a single feature to rank question-answer pairs. From the table, it is clear that this feature alone significantly outperforms the random baseline in most datasets. For Squad-sent, which is our adaptation of the SQuAD v. 2.0 dataset where the task is to identify the sentence containing the answer, this feature alone identifies the sentence containing the answer $65.48\%$ of the times. 
All of the recent models in the literature have tried to model such features; for example, \citet{severyn2016modeling} uses the relational feature that marks words appearing in both the question and the answer, and many state-of-the-art approaches have primarily used the attention mechanism \cite{wang2016compare, bian2017compare, sha2018multi}.

\subsubsection{Global Structure}
Another relevant feature for A2S datasets is the global structure present in the original rank. The structure of the document in a document-based dataset provides an important signal for answer sentence selection. Table \ref{tab:datasets} shows that in the case of WikiQA, SQuAD, and Natural Questions, there is a high chance that the answer is contained in the first sentence/paragraph. This is particularly true for WikiQA and Natural Questions. In these datasets, the P@1 computed on the original sentence rank (order of the sentences in the raw text) is $\sim 46$. There may be several reasons for this distribution. For example, we believe that there is an intrinsic correlation between the real world distribution of questions and the structure of the Wikipedia document: encyclopedic knowledge is usually organized in a way that more general information about a topic is summarized and organized at the beginning of the document. 

In contrast, the signal is less present in datasets such as SQuAD, where annotators are asked to write questions after reading the whole paragraph, i.e., they target each part of the text by construction. Thus the answer distribution is less skewed. However, for the same reason, it is important to note that annotators tend to introduce more lexical overlap bias when writing questions after reading the source of the answers.

Additionally, Table \ref{tab:datasets} shows that the combination of the two features, word-overlap, and reciprocal rank, gives a strong baseline for all the datasets in consideration. This simple rule-based model ranks candidates according to the lexical overlap between question and candidates, and, in the case when two sentences have the same amount of overlapping words, it uses the reciprocal rank (RR) as a discriminator. 

%\begin{figure*}[!t]
%\centering
%  \includegraphics[max width=\textwidth]{plots.pdf}
%  \caption{Distribution of the answer position (line number) in the considered datasets. WikiQA, SQuAD and Natural Questions have candidates one target document.  We observe a distribution more skewed towards the beginning for the first type than in the second type.}
%  \label{fig:dist-wikiqa}
%\end{figure*}

\subsection{Related Work}

Despite the importance of global structure, most state-of-the-art models \cite{severyn2016modeling, he2015multi, madabushi2018integrating, tay2018multi, TANDA} do not take the global structure of candidates into account. They use a pointwise approach to maximize the score of positive candidates, i.e., candidates that contain the answer and minimize the score of the candidates not containing the answer. Most models treat ranking as a binary classification problem. Nevertheless, other methods have been studied, e.g., a contrastive pairwise and, more recently, list-wise approaches. 

In the case of contrastive pairwise training \cite{rao2016noise}, given a question q, the loss maximizes the score of the model for a positive question candidate pair $(q, cˆ{+})$ with respect to the score of the negative pair $(q, cˆ-)$. This approach intrinsically balances the distribution of positive and negative examples in training and can improve the overall results. In particular, when paired with a hard negative sampling strategy. However, the comparison between the answer candidates is performed at the score level; therefore, the model itself remains effectively pointwise.  

The list-wise approach have been proposed in recent paper \cite{bian2017compare}: the model predicts the score for each question candidate pair individually but it applies a softmax function on top of the scores given by the model $s_1, s_2, ..., s_n = softmax(\phi(q, c_1), \phi(q, c_2), ..., \phi(q, c_n))$. This approach helps in providing stability in the training process. However, as for the contrastive pairwise approach, the underlying model remains agnostic of the global structure of the rank. 

Most A2S models have a way to identify the lexical overlap and, in particular, the semantic word-overlap between question and answer.
The first neural models developed to solve the task \cite{yu2014deep, severyn2015learning} directly added the lexical overlap feature in the model by concatenating it to the question-candidate representation or as a feature concatenated to the word embeddings of Convolutional Neural Network (CNN). Subsequent approaches, e.g., \cite{wang2016compare, bian2017compare}, use a word-level attention mechanism to identify the semantic overlap between each word in the question and each word in the answer candidate. Despite obtaining better results than previous approaches, the computational cost of performing word-level attention together with the aggregation steps to leverage the information extracted by the attention mechanism increases the computational cost with respect to previous approaches.  
More recent methods \cite{lai2019gated,TANDA,yoon2018learning} leverage large pretrained contextualized word representation, e.g., BERT, ELMo, RoBERTa, for the task. These approaches achieve state-of-the-art results for A2S, but they require significant computational power for both pre-training, fine-tuning, and testing on the final task. Additionally, they require large resources in terms of data and computational power. These may not be available for low resource languages or domain-specific applications.

\section{Efficient Model for A2S}
To build an efficient yet accurate model for A2S, it is crucial to leverage all the strong signals present in the dataset without increasing the complexity of the model itself.
For this reason, we design a model as follow:
\begin{itemize}
    \item We build an efficient encoder to capture the lexical-overlap of the question-candidate pair, i.e., our \emph{Cosinet}.
    \item We add a recursive neural network on top of the question-candidate pairs to capture the global structure in the original rank.
    \item We apply a global, list-wise, optimization approach to rank all the candidate pairs jointly.
\end{itemize}

\subsection{Cosinet}

The Cosinet has three building blocks: (i) a word-relatedness encoder that performs the cosine similarity between the word embeddings in the question and the answer (generating word relatedness features); (ii) similarly to \cite{severyn2016modeling}, the relational features are concatenated to the word embeddings and fed to one layer of CNN, to create a representation for the question and candidate pair; and (iii) similarly to \citet{chen2017enhanced}, the information of the question and the candidate is combined at classification stage, by concatenating the vectors. That is, we use the component-wise multiplication and difference between question and answer vectors.

\subsubsection{Word-Relatedness encoder}
To encode the word-relatedness information, we first map the words in the question and the answer to their respective word embeddings. We then perform a comparison between all the embeddings in the question $w_i^q$ and all the embedding of the answer $w_j^c$ using the cosine similarity. 
\begin{equation}
    r_{i,j} = {{ w_i^q} {w_j^c} \over \|{w_i^q}\| \|{w_j^c}\|} 
\end{equation}

Instead of performing the weighted sum of the embeddings as in standard attention, for each word in the question, we take the maximum relatedness score between the word embedding of the question and each word embedding of the candidate, i.e., $\textbf{r}_i = max_{j}(r_{i,j})$. The same process is performed for each word in the answer. This feature represents \emph{how much a word is similar to the most similar word in the other text}.

This simple feature is concatenated with the word embedding of the question $\hat{w}_i^q = [\hat{w}_i^q; \textbf{r}_i]$ and vice-versa $\hat{w}_j^c = [\hat{w}_j^c; \textbf{r}_j]$. The resulting word representations are passed to the question-candidate encoder to create the pair representation.

It is important to note that we keep the word embedding static during training, so this operation does not cause much overhead during training as we do not need to back-propagate through it.

For this model, we use the word embeddings, Numberbatch \cite{speer2017conceptnet}, since they are more accurate than unsupervised word embeddings, such as Glove \cite{pennington2014glove}, which may introduce noisy or non common-sense relations. In particular, \citet{faruqui2016problems} showed that unsupervised word embeddings tend to cluster according to the frequency of words of the dataset used for training them. For this reason, \citet{speer2017conceptnet} adopted retrofitting. This technique aims at reducing the distance between word embeddings of entities that are related in a knowledge base, i.e., ConcepNet.

\subsubsection{Question-Candidate encoder}
Similarly to \citet{severyn2016modeling} we encode the question and candidate independently using two single layers of CNN with a kernel size of 5 and global max pooling. This results in an embedding for the question $q_{e}$ and the candidate $c_{e}$. The two embeddings are combined using them in a pair embedding, concatenating the point-wise multiplication of the two embeddings with their difference, i.e., $qc_{e} = [q_{e} \odot c_{e}; q_{e} - c_{e}]$

\subsection{Global optimization}
Standard approaches take the pair representation $qc_e$ and apply a feed-forward network that outputs the score $s_i$ for the pair $(q, c_i)$. However, this simple model is unable to capture the inter-dependencies between the candidates for the question $q$, i.e., it does not capture the global structure of the rank.

In this paper, to leverage the global structure of the rank, we make use of a Recurrent Neural Network applied on top of the $qc_{e}$ representations for each $c_{i}$ of a given question $q$. 
The resulting contextual representations $\hat{qc}_{e}$ are passed to the feed-forward network (in our experiments, we use a single layer to produce the final score).

Finally, similarly to \citet{bian2017compare}, we apply a softmax function to the scores $s^1,..,s^n$ of all the $n$ candidate answers of a given question. Then, we minimize the KL-divergence of the predicted probabilities and the normalized gold labels. \citet{bian2017compare} proved that this approach could improve the convergence speed of the model when compared with pointwise approaches.

\section{Experiments}

\subsection{Datasets}

To align with previous work, we remove the questions without answers and, we lowercase and tokenize questions and passages\footnote{Tokenization with SpaCy v2.1 \url{https://spacy.io}}.
We used the following datasets:

\paragraph{WikiQA} Questions are randomly sampled from the Bing search engine logs. The candidate answers are the sentences that constitute the first paragraph of the related Wikipedia article. Additionally, answers are concentrated in the first part of the paragraph. 
%This behavior is best shown in the analysis presented in Figure~\ref{fig:dist-wikiqa}. The Figure shows that the probability of the answer to be in the first position in the WikiQA test set is $46.09\%$,i.e., $64.26$ MRR.

\paragraph{SQuAD-sent} For each question, the SQuAD dataset provides a paragraph and annotations for the exact answer span. To adapt SQuAD for the answer sentence selection task, we split the paragraph into sentences using the SpaCy sentence tokenizer. We infer the sentence labels from the answer span labels, i.e., if a sentence contains the answer span is labeled a positive example, negative otherwise. Since the SQuAD test set is not publicly available, we use the validation set for testing and 10\% of the questions in training set for validation. In contrast with our dataset, QNLI (the GLUE adaptation of SQuAD \cite{wang2018glue}) provides an even amount of positive and negative question/answer pairs, sampled from the SQuAD dataset. This creates decontextualized sentences, which prevent to exploit the sequential structure.
In contrast, we propose a dataset that maintains the original document structure. 
We evaluate our models using two metrics:  Precision at 1 (P@1) and Mean Reciprocal Rank (MRR). The SQuAD dataset exhibits a much more substantial lexical overlap between question and answer passages. This effect can be noted in Tab.~\ref{tab:datasets}: the simple word-overlaps count baseline, i.e., the number of unique words that appear in both question and passage, achieves a P@1 of $65.48$.  

%\subsubsection{MSMarco-rank} Similarly to the WikiQA dataset, MSMarco is composed of questions sampled from the Bing search engine logs. Questions are paired with the 10-best passages from the search engine itself. When building the dataset, annotators are asked to write an answer to the question and to select the passages supporting the given answer. In this paper, we use the original MSMarco dataset for re-ranking the Bing-retrieved passages. This step aims to rank the passages that support the answer of the annotator higher than all the other passages. In this work, we do not focus on answer generation (or extraction); thus, we evaluate the MSMarco dataset with standard retrieval metrics. Differently from the others, the MSMarco is an IR dataset, where each passage is retrieved. Thus, the natural answer distribution is less skewed, e.g., there is a smaller percentage of answers in the first position (P@1  of $14.75\%$ and a MAP of $34.65$). The annotations for the test set of MSMarco are not publicly available. Therefore, we used the official development set for testing and a portion of the training set for validation. 
\begin{table}[!t]
\begin{center}
\resizebox{0.97\columnwidth}{!}{
\begin{tabular}{lll}
\hline \bf Model & \bf MAP & \bf MRR \\\hline\hline
\multicolumn{3}{c}{Baselines}\\\hline
RR & 64.21 & 64.26\\
WO  &51.02 & 51.24 \\
WO+RR & 68.25 & 69.43 \\

\hline\hline
\multicolumn{3}{c}{Related Work w/o pre training}\\
\hline
\citet{tay2018hyperbolic}  & 71.20 & 72.70 \\
$\mathrm{W\&J}$ \citeyear{wang2016compare}   & 74.33 & 75.45 \\
$\mathrm{W\&J}$ \citeyear{wang2016compare}$\dagger$ & $72.38 \pm 1.4$ & $73.44 \pm 1.5$  \\
\citet{sha2018multi} & 74.62& 75.76\\
\citet{bian2017compare}  & \textbf{75.40} & \textbf{76.40} \\\hline
\multicolumn{3}{c}{Related Word with pre-training}\\
\hline
\citet{yoon2018learning}  &83.40  & 84.80  \\
\citet{lai2019gated} &85.70  & 87.20  \\
\citet{TANDA}   &92.00  & 93.30  \\\hline\hline
\end{tabular}}
\end{center}
\caption{\label{tab:wikiqa} Related Work on WikiQA test-set. $\dagger$We run the official implementation with different random seeds.}

\end{table}

\begin{table*}[!t]
\begin{center}
\resizebox{0.74\linewidth}{!}{
\begin{tabular}{llrrrr}
\hline \bf Model  & \textbf{RNN} &\bf MAP & \bf MRR & \bf params & \bf train-time \\ \hline\hline
\multicolumn{6}{c}{Baselines}\\\hline
RR & - & 64.21  & 64.26 & -  & - \\
WO  & - &51.02 & 51.24& - & - \\
WO + RR & - &68.25 & 69.43& - &  -\\

\hline\hline
\multicolumn{6}{c}{Our Models}\\\hline
$\mathrm{Cosinet}$  & -& $70.95\pm0.6$&    $72.86\pm0.7$   & 904k & 6 sec \\
$\mathrm{Cosinet}_{list}$ & - &  $71.22\pm0.2$&$73.07\pm0.3$  & 904k & 5.5 sec\\
$\mathrm{Cosinet}_{list}$ & RNN &  $74.78\pm0.6$&$76.35\pm0.6$  &  1.17M & 7.5 sec \\
$\mathrm{Cosinet}_{list}$ & BiRNN &  $\mathbf{75.62}\pm0.8$&$\mathbf{77.13}\pm0.9$ & 1.12M & 8.9 sec \\
$\mathrm{Cosinet}_{list}$ & LSTM &  $74.31\pm0.8$&$75.78\pm0.9$  & 1.99M & 7 sec\\
$\mathrm{Cosinet}_{list}$ & BiLSTM &  $75.32\pm0.6$&$76.85\pm0.5$ & 1.81M & 9.5 sec \\
\hline
$\mathrm{CA}$  & - & $72.03\pm1.6$&$73.39\pm1.7$  & 2.89M & 19 sec \\
$\mathrm{CA}_{list}$ & - &  $71.43\pm1.0$&$73.55\pm1.0$ & 2.89M & 18 sec\\
$\mathrm{CA}_{list}$ & RNN &  $74.73\pm1.0$&$76.35\pm1.2$ &  5.05M & 20 sec \\
$\mathrm{CA}_{list}$ & BiRNN & $74.97\pm1.2$&$76.44\pm1.2$ & 4.87M & 21 sec \\
$\mathrm{CA}_{list}$ & LSTM &  $74.82\pm1.1$&$76.42\pm1.2$ & 11.53M & 25 sec\\
$\mathrm{CA}_{list}$ & BiLSTM &  $74.27\pm1.0$&$75.74\pm1.1$  & 10.81M & 25 sec \\
\hline
$\mathrm{BERT}_{base}$ & - & $81.32$&  $82.50$ & 110.00M & 17 min 50 sec \\
\hline
\end{tabular}}
\end{center}
\caption{\label{WikiQA} Model comparison on the WikiQA test-set.}
\end{table*}

\paragraph{NQ-LA} The Natural Question dataset uses question sampled from the Google search engine logs. The questions are given to the annotators together with the retrieved Wikipedia page. The annotator is asked to select (i) a long answer, i.e., the smallest HTML bounding box containing all the information needed to answer the question, and (ii) a short answer (if available), that is, the actual answer to the question.
We consider only paragraphs as long answers, removing tables and lists. As the latter requires a different semantic approach than the one typically used for free text. A paragraph is defined by the HTML bounding box \textless p\textgreater and \textless \textbackslash p\textgreater. 
The dataset has a similar answer distribution of the others, i.e.,  P@1 $46.06\%$ and MAP $57.30$, even if the candidates are much longer (paragraphs). These results are interesting considering that a Wikipedia page contains an average of 30.95 paragraphs (of $98.76$ words). We note that most pages give essential information about an entity in the first paragraph, i.e., in the summary paragraph.
Similarly to SQuAD, the annotations for the test set of Natural Questions are not publicly available. Therefore, we used the official development set as our test set and a portion of the training set for validation.

\begin{table*}[!h]

\begin{center}
\resizebox{0.69\linewidth}{!}{
\begin{tabular}{llrrrr}
\hline \bf Model  & \textbf{RNN} &\bf P@1 & \bf MRR & \bf params & \bf train-time \\ \hline\hline
\multicolumn{6}{c}{Baselines}\\\hline
RR &-& 30.55  & 53.53&-&-\\
WO &- &65.48 & 77.90&-& -\\
WO + RR &-&73.12 & 83.60&-& -\\

\hline\hline
\multicolumn{6}{c}{Our Models}\\\hline
$\mathrm{Cosinet}$  && $86.18\pm0.2$&$91.81\pm0.1$   & 904k & 1 min 47 sec \\
$\mathrm{Cosinet}_{list}$ &&  $85.12\pm0.1$&$91.16\pm0.1$  & 904k & 8 min 10 sec\\
$\mathrm{Cosinet}_{list}$ & BiRNN & $86.18\pm0.2$&$91.97\pm0.1$ & 1.12M & 12 min 30 sec \\
\hline
CA  &&$85.71\pm0.2$&$91.49\pm0.1$& 2.89M & 6min 30 sec \\
$\mathrm{CA}_{list}$ &&$85.17\pm0.6$&$90.69\pm1.0$& 2.89M & 24 min 11 sec\\
$\mathrm{CA}_{list}$ & BiRNN & $\textbf{86.32}\pm0.3$&$\textbf{92.05}\pm0.2$& 4.87M & 28 min 30 sec\\
\hline
$\mathrm{BERT}_{base}$ & - & $92.44$ & $95.62$& 110.00M & 6 hr 50 min \\
\hline
\end{tabular}}
\end{center}
\caption{\label{SQuAD} Model comparison on the SQuAD sent test-set.}
\end{table*}

\subsection{Models and parameters}

In our experiments, we used two different encoder architectures: the newly proposed Cosinet and our re-implementation of the Compare-Aggregate (CA) architecture. The former uses static Numberbatch embeddings\footnote{https://github.com/commonsense/conceptnet-numberbatch} of size 300; the convolution hidden layer of size 300 and a kernel of size 5. For the CA architecture, we use the standard parameters of the original paper, but in contrast with it, we keep the embedding static as we empirically found that it leads to similar results while having the highest number of trainable parameters.
For the RNN and the LSTM, we used the same hidden size as the input, i.e., double the size of the convolutional operation hidden size. For the Bidirectional variations, i.e., BiRNN and BiLSTM, we set the hidden size as half of the input size in each direction, resulting in a comparable number of parameters.

All the models were trained for three epoch using slanted triangular learning rate scheduling \cite{howard2018universal} without early stopping. In the case of the pointwise models, we used Adam optimizer with a maximum learning rate set at 2e-3, whereas for the list-wise approaches, we used a learning rate of 2e-4.
All the experiments are performed on an Nvidia GTX 1080 ti GPU and an Intel Core I9-7900X processor.

\subsection{Results}
\begin{table*}[!t]
\begin{center}
\resizebox{0.62\linewidth}{!}{
\begin{tabular}{llrrrr}
\hline \bf Model  & \textbf{RNN} &\bf MAP & \bf MRR & \bf params & \bf train-time \\ \hline\hline
\multicolumn{6}{c}{Baselines}\\\hline
RR & - & 57.30  & 60.41& - & -\\
WO & - &38.09 & 39.57&- & -\\
WO + RR &- & 53.98& 56.45& -&- \\
\hline\hline
\multicolumn{6}{c}{Our Models}\\\hline
$\mathrm{Cosinet}$  && 69.74 & 72.69  & 904k & 1 hr 13 min \\
$\mathrm{Cosinet}_{list}$ && 68.16 & 71.06 & 904k & 17 min\\
$\mathrm{Cosinet}_{list}$ & BiRNN & 73.28 & 76.05& 1.12M & 34 min \\
\hline
$\mathrm{CA}$  &&69.88&72.77& 2.89M & 5h 39min \\
$\mathrm{CA}_{list}$ &&69.82&72.78& 2.89M & 2h \\
$\mathrm{CA}_{list}$ & BiRNN & \textbf{74.21} &\textbf{76.88}& 4.87M & 2h 10 min\\
\hline
\end{tabular}}
\end{center}
\caption{\label{NQLA} Model comparison on the NQ-LA test-set.}
\end{table*}

Table \ref{tab:wikiqa} shows the results in the WikiQA dataset. The first block reports the performance of the baselines; these models are computed using the simple features described in Section \ref{sec:datasets}, i.e., the lexical overlap and the reciprocal rank: both achieve results comparable with baseline CNN architectures.  The second block of results shows the performance of models from previous work that do not use pretrained language models. The third section presents the results of models that use both pretrained language models and transfer learning. In particular, \citet{yoon2018learning} uses the ELMo and transfer learning on QNLI dataset; \citet{lai2019gated} uses BERT and performs transfer learning on the QNLI dataset, and \citet{TANDA} uses RoBERTa large and performs transfer learning from he Natural Question dataset.
The performance of our approach is shown in table \ref{WikiQA}: the Cosinet architecture achieves comparable results with respect to the more complex CA while having much lower trainable parameters. Cosinet and CA are the standard pointwise approach, trained on all the data using a fixed batch size and binary cross-entropy (BCE). $Cosinet_{list}$ and $CA_{list}$ are the same base architecture but trained with a listwise approach, with KL-Divergence loss on all the question-answer candidate pairs for the same question. We then analyzed what RNN architecture is best suited to identify the structure of the original rank. For both Cosinet and CA, we found that there is no much statistical difference between RNN and LSTM.  However, Bidirectional RNNs seems to outperform the other models consistently. %In the last block, we present our results with $BERT_{base}$. %As expected, BERT outperform all the proposed approaches, but it requires much more parameters and longer training times

The same is true for the SQuAD dataset in Table \ref{SQuAD}. Our base architecture achieves comparable results with the more complex CA model. Moreover, adding the BiRNN to the model further improves the results.
It is important to note how simple semantic matching models are capable of achieving very high results on the task. In SQuAD, the lexical overlap feature is more prominent than the global rank feature. Therefore the major impact is given by the Global Optimization with the BiRNN.

Finally, Table \ref{NQLA} shows the results of the bigger Natural Question dataset. Despite the different nature of the data, i.e., the candidates are paragraphs rather than sentences, our proposed model improves with respect to the baselines in particular when combined with the BiRNN.

\subsection{Performance analysis}

From the experiments on all three datasets is clear that the proposed architecture is more efficient than the CA one since it has much fewer parameters and a more efficient attention computation. On the small WikiQA dataset, the model takes up to 9.5 seconds to train, achieving results that are comparable with the best models that do not make use of pretrained language models — in contrast, training BERT-base on the same dataset requires 17 min and 50 sec. Additionally, our model has roughly 100x less trainable parameters than BERT-base. The difference between the two models is more evident when comparing the training time on the SQuAD dataset. The BiRNN based Cosinet trains in slightly more than 12 minutes, which is much lower than the 6 hours and 50 minutes needed to train BERT-base on the same task, and around half the time needed to train the CA architecture.

\section{Conclusions}

In this paper, we argue that by exploiting the intrinsic structure of the original rank together with an effective word-relatedness encoder, we can achieve competitive results with respect to the state of the art while retaining high efficiency.  We first analyzed the structure of standard datasets, highlighting the importance of the global structure in the original rank. Capitalizing on this, we propose a model that both exploits the rank structure using a simple RNN and the standard word-relatedness features, while preserving high efficiency. The model uses around 1M parameters depending on the configuration and achieves better results than the previous work that does not make use of computationally expensive pre-trained language models. Our model takes 9.5 seconds to train on the WikiQA dataset, i.e., very fast in comparison with the $\sim 18$ minutes required by a standard BERT-base fine-tuning.  On SQuAD, this difference is even higher, i.e., minutes vs. hours, required by Transformer-based models.

\bibliography{acl2020}
\bibliographystyle{acl_natbib}

\end{document}